\documentclass[conference]{IEEEtran}
\IEEEoverridecommandlockouts
\usepackage{cite}
\usepackage{amsmath,amssymb,amsfonts}
\usepackage{algorithmic}
\usepackage{graphicx}
\usepackage{textcomp}
\usepackage[table]{xcolor}
\def\BibTeX{{\rm B\kern-.05em{\sc i\kern-.025em b}\kern-.08em
    T\kern-.1667em\lower.7ex\hbox{E}\kern-.125emX}}

\newcommand\T{\rule{0pt}{2.9ex}}       % Top strut
\newcommand\B{\rule[-1.2ex]{0pt}{0pt}} % Bottom strut 
\usepackage{caption}
\usepackage{wrapfig}
\usepackage{tikz}
\usepackage{subcaption}
\usepackage[nobiblatex]{xurl} %Keeps URL within margins on the references
\usepackage{hyperref} 

\hypersetup{
    colorlinks=true,
    linkcolor=red,
    urlcolor=blue,
    linktoc=all,
    citecolor=green}
\usepackage{csquotes} %Allows for speech marks 

\begin{document}

\title{Identification of Driver Phone Usage Violations via State-of-the-Art Object Detection with Tracking\\

}

\author{\IEEEauthorblockN{Steven Carrell}
\IEEEauthorblockA{\textit{School of Computing} \\
\textit{Newcastle University}\\
Newcastle upon Tyne, United Kingdom \\
steven.carrell.ncl@gmail.com}

\and
\IEEEauthorblockN{Amir Atapour-Abarghouei}
\IEEEauthorblockA{\textit{Department of Computer Science} \\
\textit{Durham University}\\
Durham, United Kingdom \\
amir.atapour-abarghouei@durham.ac.uk}
}

\maketitle

\begin{abstract}
The use of mobiles phones when driving have been a major factor when it comes to road traffic incidents and the process of capturing such violations can be a laborious task. Advancements in both modern object detection frameworks and high-performance hardware has paved the way for a more automated approach when it comes to video surveillance. In this work, we propose a custom-trained state-of-the-art object detector to work with roadside cameras to capture driver phone usage without the need for human intervention. The proposed approach also addresses the issues caused by windscreen glare and introduces the steps required to remedy this. Twelve pre-trained models are fine-tuned with our custom dataset using four popular object detection methods: YOLO, SSD, Faster R-CNN, and CenterNet. Out of all the object detectors tested, the YOLO yields the highest accuracy levels of up to $\sim$96\% ($AP_{10}$) and frame rates of up to $\sim$30 FPS. DeepSort object tracking algorithm is also integrated into the best-performing model to collect records of only the unique violations, and enable the proposed approach to count the number of vehicles. The proposed automated system will collect the output images of the identified violations, timestamps of each violation, and total vehicle count. Data can be accessed via a purpose-built user interface.
\end{abstract}

\begin{IEEEkeywords}
Mobile phone detection, YOLO Object Detection, Intelligent Transportation Systems, Deep Learning
\end{IEEEkeywords}

\section{Introduction}
According to  the World Health Organization (WHO), approximately 1.3 million people die each year as a result of road traffic accidents \cite{r1}. A contributing factor towards this is the use of a handheld mobile devices while operating a motor vehicle. Drivers using a mobile phone are approximately four times more likely to be involved in a crash than drivers not using their phone \cite{r1}. 

In 2017, the UK Government doubled the penalty for using a mobile phone while driving to 6 points and a £200 fine (up from 3 points and £100) \cite{r3}. A study carried out in 2015 suggests that there is a negative correlation between a higher fine and the likelihood of a person using their phone \cite{r4}.

Typically, catching drivers using their phones involves road-side police performing the laborious task of capturing the violation as it happens. Unless there are significant resources dedicated to this task, there is a strong probability that many of these violations will go undetected. This opens the door for a more automated process of capturing these violations.

Recent years have seen the development of object detection in video surveillance \cite{r41}. Object detection frameworks such as Faster Region Based Convolutional Neural Networks (Faster R-CNN) \cite{r19} and Single Shot Detector (SSD) \cite{r23} have made it possible to take video images and detect objects with both high accuracy and speed \cite{r5}.

In this work, we propose a fully-automated system that will take live video from roadside surveillance cameras and detect if a driver is using a mobile phone whilst the vehicle is in operation. We will explore different quality cameras (high-end and low-end) whilst addressing challenges such as windscreen glare, tinted windows and low-light scenarios. In order for the system to be fully automated, it will need to have the ability to log each unique violation as well as to save the images corresponding to each violation.

\begin{figure}[!tbp]
  \centering
%   \subfloat[Step one.]{\includegraphics[width=3.5cm, height=3cm]{two_step2.0.JPG}}
  \subfloat[Step one.]{\includegraphics[width=4.2cm, height=3cm]{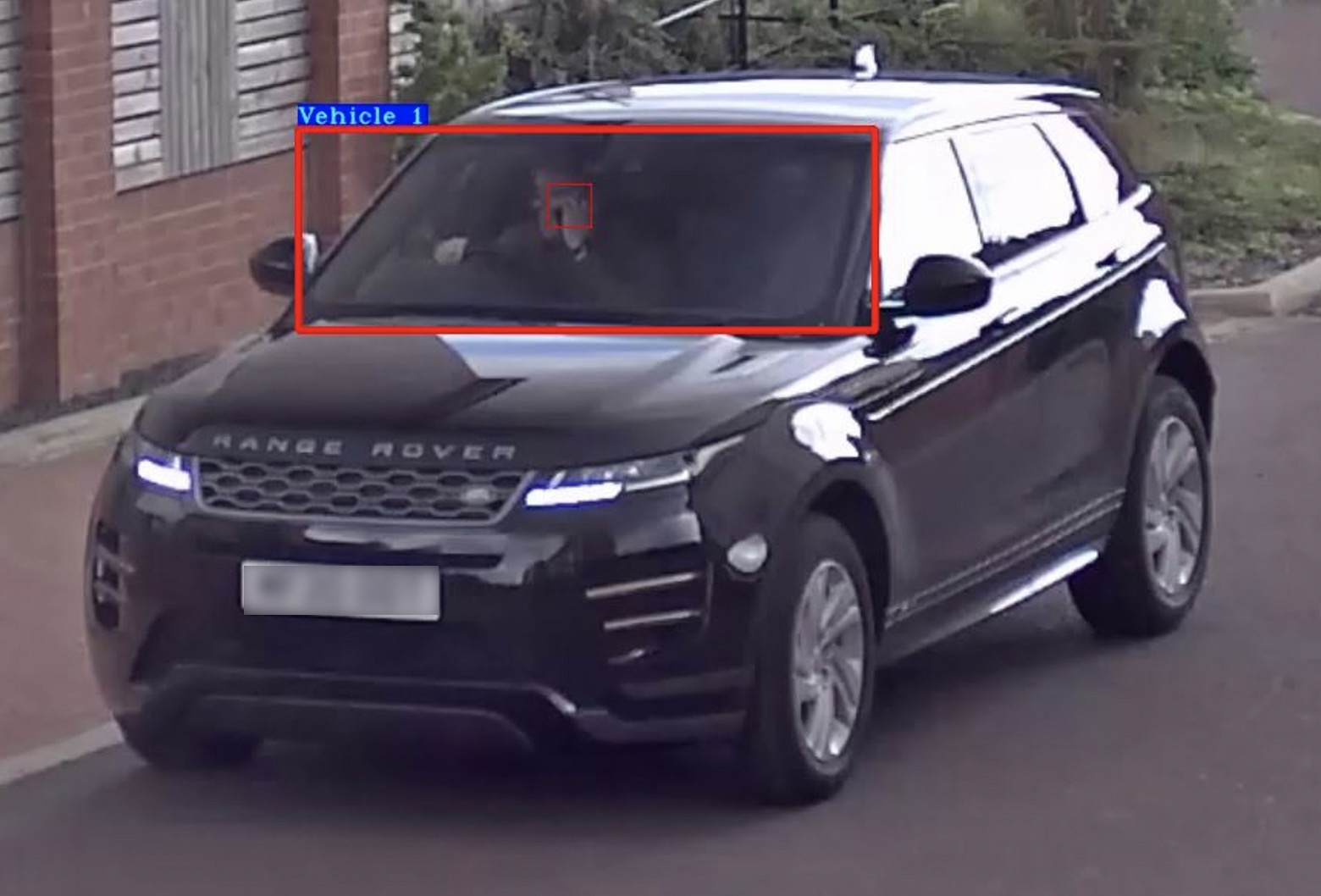}}
  \hskip 1ex
  \subfloat[Step two.]{\includegraphics[width=4cm, height=3cm]{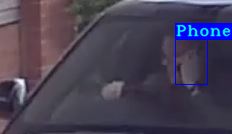}}
  \caption{Example of the proposed two-step approach: the first step (left) detects windscreen; the second step (right) first crops the driver's side and only then detects the phone.}
  \label{fig:two_step}\vspace{-0.5cm}
\end{figure}

We propose two methods for achieving this task; first, a single-step method focusing on efficiency and speed, using a single trained model to detect violations. This method will suffer a trade-off with accuracy due to potential issues caused by having to find an extremely small object (phone) within a large image. The second method (two-step) focuses on achieving high accuracy by running two individually trained models simultaneously.

The single-step system is trained to detect two classes, namely mobile phone and licence plate with the plate only used as a method of counting the total number of vehicles. The two-step system (Figure \ref{fig:two_step}) first detects the windscreen and then uses the cropped image of the driver's side of the windscreen as the input to the second step to detect the mobile phone. Similar to the licence plate in the first (single-step) method, the windscreen is used to count the number of vehicles so the number of detections can be worked out as a proportion to the total number of vehicles. Models are trained using a number of different state-of-the art object detector frameworks. These include YOLO (You Only Look Once) v3 \cite{r9} and v4 \cite{r10}, SSD \cite{r20, r21}, Faster R-CNN \cite{r19} and Centernet \cite{r21}. 

%  \begin{figure*}[htp]
%     \centering
%     \includegraphics[width=.9\textwidth]{flow.JPG}
%     \caption{FPS for detection only for the trained object detectors.}
%     \label{fig:flow}
% \end{figure*}

Model performance is measured via Average Precision (AP) \cite{r30}, and using both the PASCAL VOC evaluation metric where the Intersection Over Union (IoU) score is $>$0.5  \cite{r11}, whilst also testing IoU$>$0.1 due to the nature of objects being detected in this particular application. The proposed solution will be designed to work with a live video; therefore, we also evaluate the efficiency of the proposed method by calculating the frame rate of the output - i.e. frames per second (FPS). The images that are used to train our model on the phone class will predominantly be obtained and created especially for this project. In order for the model to detect mobile phone use violations with a reasonable level of accuracy, the training images will be replications of the real-world scenario of a person using their phone whilst driving (examples of training images can be seen in Figure \ref{fig:images1}).

Once the final model is chosen, we integrate a tracking algorithm \cite{r8} into our model to avoid the issue of logging multiple detections for the same violation. This will allow the system to keep track of the total number of violations for a given duration. This data can then be analysed using a purposely designed user interface.

This work aims to successfully develop a system that can work with a roadside camera 24 hours a day to automatically detect whether a person is using their mobile phone when in operation of a motor vehicle. In short, the primary contributions of this paper are as follows:
\begin{itemize}
    \item Train and evaluate multiple object detection methods \cite{r9, r10, r20, r21, r19} to detect mobile phone use violations with a reasonable degree of accuracy and speed by establishing an appropriate trade-off between predictive performance and efficiency.
    \item Test the trained models on both full images and cropped windscreens to determine if a single-step or two-step approach is more appropriate. The single-step method operates on the full image only, while the two-step system first finds windscreen and then uses this cropped image to detect the phone (Section \ref{sec:appraoch}).
    \item Establish what can be achieved with on a low-cost budget using a low-end consumer camera solution as well as providing insights on what can be achieved with a reasonable budget using high-end cameras.
    \item Ensure issues such as windscreen glare and poorly-lit environments are addressed so the system can work at any time of the day.
    \item Integrate a tracking algorithm \cite{r8} to identify unique detections in order to log useful data whilst providing snapshots of the violations. Collected data can be accessed through a purposely designed user interface.
\end{itemize}

To better enable reproducibility, the source code for the project is publicly available\footnote{\url{https://github.com/carrell-ncl/Windscreen2}}.

The remainder of the paper is organised as follows. Related work on mobile phone usage detection within the existing literature is presented in Section \ref{sec:related_work}. The proposed approach is described in Section \ref{sec:appraoch}. Experimental results are examined and discussed in Section \ref{sec:results}. We address limitations and future work in Section \ref{sec:discussions}, before finally concluding the paper in Section \ref{sec:conclusion}.

\section{Related Work}
\label{sec:related_work}

We consider related work in the context of object detection in general (Section \ref{sec:related_work:object_detection}) and specific systems utilised in distracted driver identification applications (Section \ref{sec:related_work:distracted_driver}).

\subsection{Object Detection}
\label{sec:related_work:object_detection}

Traditionally, any kind of object identification with the use of video surveillance would be done manually and could involve the laborious task of humans either monitoring the live stream of the camera or reviewing historic footage. This difficult and tiresome task meant that a more intelligent approach would eventually be required \cite{r43}. Object detection is a computer vision task concerned with detecting and classifying objects within an image. The use of this technology has paved the way for a more automated solution into video surveillance applications. Examples of such tasks include Licence Plate Recognition (LPR) \cite{r32}, people tracking, vehicle counting, or unattended baggage in airports.

We can describe two-major historic milestones in the development of object detection: \enquote{traditional object detection period} (pre 2014) \cite{r45, r46, r47} and \enquote{learning-based detection period} \cite{r44}. It was not until 2015 where object detection could be utilised in real-time video with the development of Faster Regional-Based Neural Network (Faster R-CNN) \cite{r19}, an improvement from its predecessor R-CNN \cite{r49} in both accuracy and speed. 

Modern-day object detectors can be categorised into two types: one-stage (YOLO \cite{r9, r10}, SSD \cite{r20}, and CenterNet \cite{r21}) and two-stage (R-CNN series including Fast R-CNN \cite{r48}, Faster R-CNN \cite{r19}, R-FCN \cite{r50},
and Libra R-CNN \cite{r51}). Two-stage detectors generally split the overall object detection task as follows: the first stage is to generate proposals, and then the second stage focuses on the verification and recognition of these proposals \cite{r35}. The two-stage detectors are typically slower due to their heavy-head design but detect with a higher level of accuracy \cite{r35}, whilst one-stage approaches tend to be faster but can be more limited in terms of their predictive performance \cite{r36}. 

\subsection{Distracted Driver Identification}
\label{sec:related_work:distracted_driver}

Recent years have seen a rise in the number of distracted driver identification systems. One such approach \cite{r6} utilises current infrastructure using LPR \cite{r32} roadside cameras. The system contains a three-stage approach to detecting whether the driver is using their phone. The first stage is to detect the windscreen, which is then cropped and processed in the second model to identify whether a person can be clearly seen. This process is in place to ensure that the images with undesirable reflection effects are not processed. The final stage will then detect for mobile phone usage \cite{r6}. The authors recognise a limitation of this system \cite{r6}; the images could not be acquired during summer days between 12:00 and 15:00 due to excessive amount of windscreen glare. Initial testing for this work shows windscreen glare for the majority of the day (as discussed in Section \ref{sec:approach:hardware:glare}), suggesting that this would not be appropriate for some territories. \\
 
 \begin{figure}[htp]
    \centering
    \includegraphics[width=9cm, height=2.5cm]{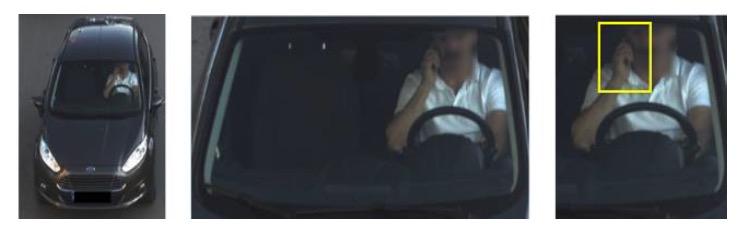}
    \caption{The three-stage mobile usage violation detection approach of Alkan et al. \cite{r6}.}
    \label{fig:Alkan}
\end{figure}
 
Another study builds upon a software already developed by the Dutch Police, which first looks for a licence plate, based on which it detects and outputs the driver's side of the windscreen \cite{r13}. These images are used as inputs to the trained model where hands, phone and face are detected. The image of the hand (taken from bounding box) is classified using VGG-16 \cite{r12}. The model looks to eliminate the issue of falsely classifying objects such as phone mounts. This is done by checking where the phone is positioned in relation to the head and hand: if the distance is greater than a set threshold, then it is not classified. The system is dependent on the Dutch Police windscreen detector, which only works for Dutch Plates. Due to the reliance on a third party software to make this work, there is a lack of control on a significant portion of the overall approach. There could be issues with support further down the line, or changes to licensing. It seems that more work would be required in order for this system to be deployable.

 Another work monitors the driver using their phone in addition to hand position \cite{r7}. This approach uses cameras inside the vehicle positioned towards both the driver and the steering wheel. They propose a Multiple Scale Faster R-CNN \cite{r19} to detect both mobile phone and hands. Geometric information is then extracted to determine if the driver is using their phone \cite{r7}. Mass deployment of this system could prove costly and impractical due to its reliance on cameras within the vehicle as well as the drivers being aware that they are being monitored.
 
 Our proposed system looks to solve the limitations of prior work starting with the issues of windscreen glare where the use of a polarising filter on the camera lens has been explored. The trained object detector is designed to work off a standard video surveillance roadside camera  and as a result, limit the cost of deployment. The next section describes our approach for building this system in more detail.

\begin{table}[h]
\centering
\begin{tabular}{ p{2.2cm}p{3.2cm}p{2.1cm} }
\hline
\rowcolor{lightgray}Object Detector & Backbone & Image Resolution\T\B \\
\hline
YOLOv3 \cite{r4} & Darknet-53 \cite{r4} & 320, 416, 512\T  \\
YOLOv4 \cite{r10}& CSPDarknet-53 \cite{r10} & 320, 416, 512  \\
Faster R-CNN \cite{r19} & Resnet101 \cite{r25} & 640 \\
Faster R-CNN \cite{r19} & Resnet152 \cite{r25}& 640  \\
Centernet \cite{r21} & Resnet101 FPN \cite{r25} & 512\\
SSD \cite{r20}& Mobilenetv2 FPNLite \cite{r27} & 640\\
SSD \cite{r20}& ResNet50 V1 FPN \cite{r25} & 640\\
SSD \cite{r20}& ResNet101 V1 FPN \cite{r25} & 640\B \\
\hline
\end{tabular}
\captionsetup[table]{skip=7pt}
\captionof{table}{Chosen object detectors and pre-trained base models (Backbone) fine-tuned and evaluated as part of this work.}
\label{tab:objdet}
\end{table}

\section{Approach}
\label{sec:appraoch}

 \begin{figure*}[htp]
    \centering
    \includegraphics[width=.9\textwidth]{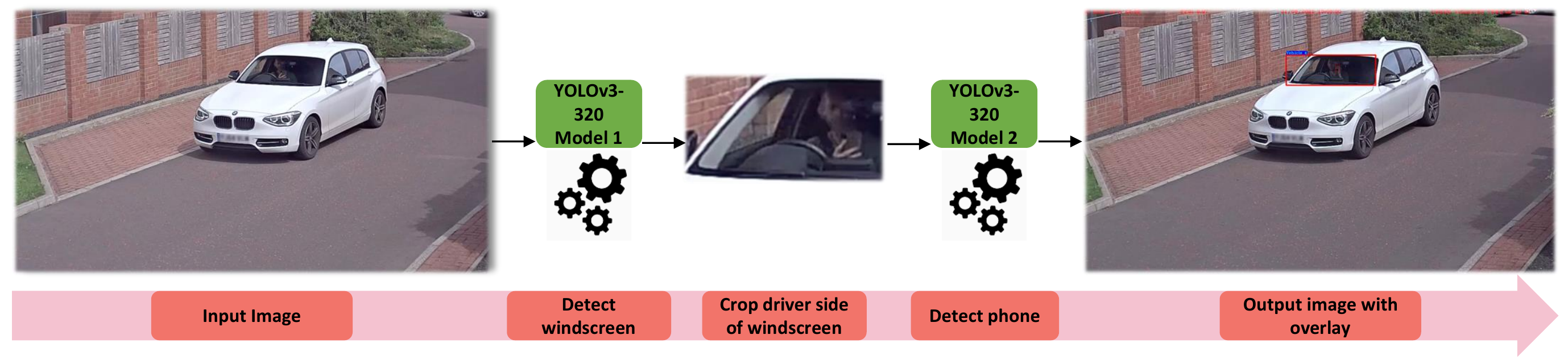}
    \caption{Two-step approach using YOLOv3 with input size of 320$\times$320. Each input frame resized to 320$\times$320 then passed through the first model to detect windscreen. Image cropped on driver side of the windscreen is then resized to 320$\times$320. Cropped image is passed through second model to detect phone. Output image is the original frame with overlay of predicted windscreen and phone bounding boxes.}
    \label{fig:flow}\vspace{-0.5cm}
\end{figure*}

Here, we describe the steps taken to build the proposed solution of a fully-automated system to detect driver violations. We propose two methods for achieving this: a single-step and a two-step approach (Figure \ref{fig:flow}), where by step we refer to a dedicated trained model in the overall system architecture. The single-step model is trained to detect both a licence plate and a person using their phone from a single image input in one forward pass through the model. A key advantage of this approach is that running a single model to complete the entire task at once results in a more light-weight faster system. A potential limitation of this method, however, would be the trade-off with accuracy as the trained model will be attempting to detect a very small object within a large image.

To remedy this issue, we also propose a two-step solution, which first detects the windscreen of the vehicle, and then uses the cropped image of only the driver side as the input for the next step to search for the phone. An overview of the process of the two-step approach is seen in Figure \ref{fig:flow}.

To enable a rigorous analysis and provide insight into the requirements of such an automated distracted driver identifications system, both the proposed one-step and two-step models will be evaluated for accuracy and speed.

To decide upon the best object detection model, we evaluate four popular frameworks with various backbone models and image input sizes \cite{r4, r10, r19, r20}. In total, we fine-tune 12 pre-trained models with our custom dataset, where we have phone and licence plate for the single-step method, and windscreen and phone for the two-step one. The chosen architectures used in this project are listed in Table \ref{tab:objdet}.

\begin{figure}
  \begin{subfigure}[t]{.5\textwidth}
    \centering
    \includegraphics[width=3.5cm, height=3.5cm]{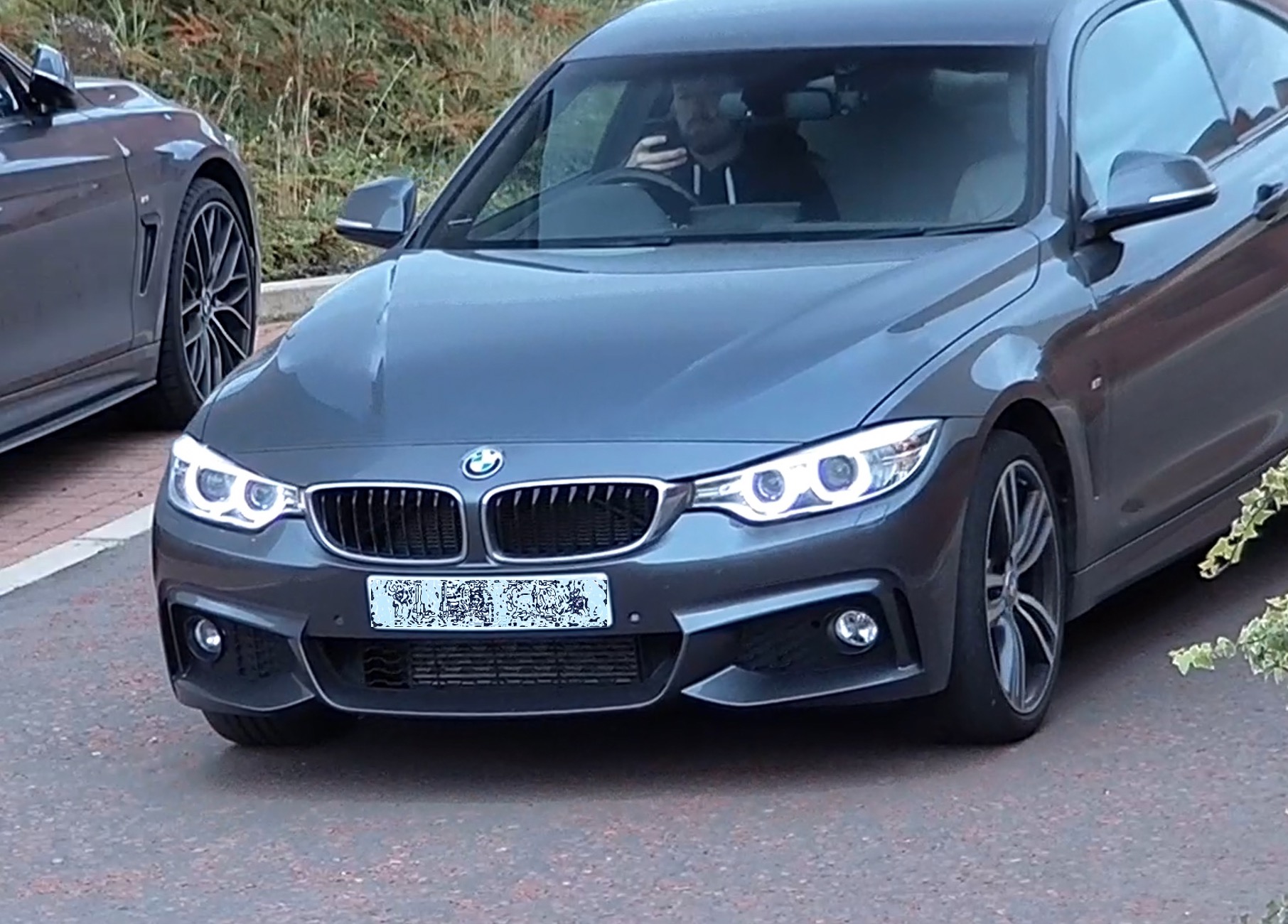}
    \includegraphics[width=3.5cm, height=3.5cm]{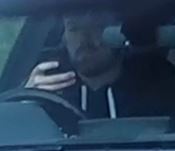}
    \caption{High-end camera example.}
  \end{subfigure}
  \hfill
  \begin{subfigure}[t]{.5\textwidth}
    \centering
    \includegraphics[width=3.5cm, height=3.5cm]{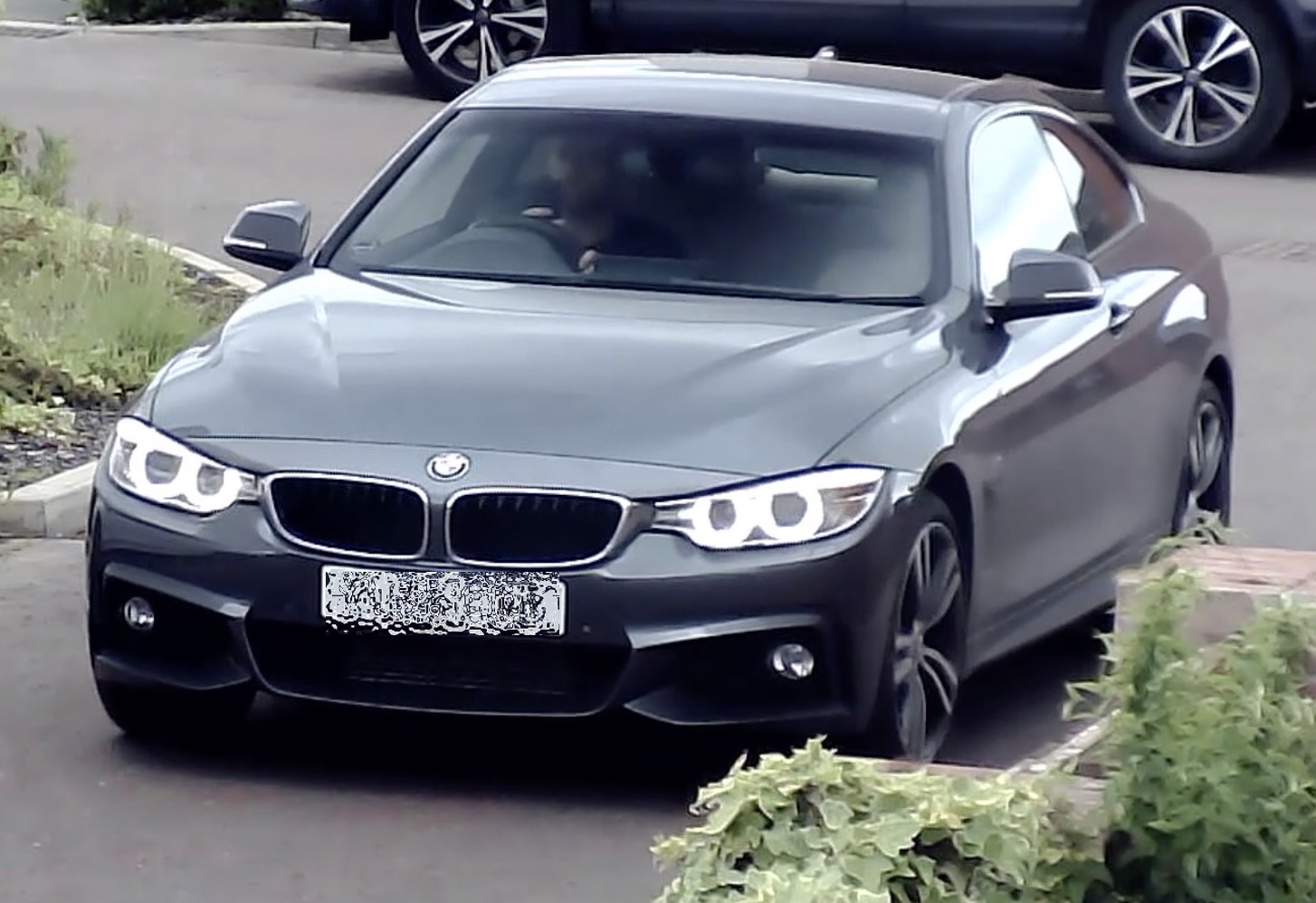}
    \includegraphics[width=3.5cm, height=3.5cm]{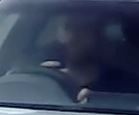}
    \caption{Low-end camera example.}
  \end{subfigure}
      \caption{Two images taken under the same conditions, using both high-end and low-end cameras.}
    \label{fig:highlow}\vspace{-0.5cm}
\end{figure}

To train and evaluate the models, images are acquired using high-end (Aviglon\footnote{\url{https://www.avigilon.com/}}, Axis\footnote{\url{https://www.axis.com/en-gb}}) and low-end (ELP\footnote{\url{http://www.elpcctv.com/}}) cameras under varying weather conditions. Details of all equipment used can be found in Table \ref{tab:equipment}. Figure \ref{fig:highlow} demonstrates the difference in quality between a high and low-end camera, where all other conditions are identical. We discuss the camera and hardware setup in the following section.

\subsection{Camera and hardware setup}
\label{sec:approach:hardware}

To successfully develop an automated mobile phone use detection system, a practical feasibility study is required to ensure that images of a high enough quality could be captured through a car windscreen in all weather conditions. It is highlighted in the work of Alkan et al. \cite{r6} that images captured in certain hours of the day present the challenge of windscreen glare, whilst night-time and poorly-lit areas can result in dark unusable images. We also acknowledge that tinted windscreens may result in cameras not having reasonable visibility into the vehicle. While this can simply be resolved by having the camera in night-mode and using excessive amounts of IR, this issue is not common in the UK due to legal restrictions, so it is beyond the scope of this work. 

\begin{table*}[t]
\centering
\begin{tabular}{ p{2.0cm}p{2.0cm}p{3cm}p{2.8cm}p{2.4cm}p{1.7cm} }
\hline
\rowcolor{lightgray}Equipment & Make & Model number & Resolution/Wavelength & Lens & Origin\T\B  \\
\hline
Camera & Avigilon & 2.0C-H5A-B1 & 2MP & 4.7 - 84.6mm & Canada  \T  \\
Camera & Axis & P1353 & 1.3MP & 5-50mm & Sweden \\
Camera & ELP & ELP-USB-FHD01M-SFV & 2MP & 5-50mm & China \\
Infra-Red & Raytec & VAR2-i8-1 & 850nm & 10 Degrees & UK \\
Infra-Red & Raytec & VAR2-i8-1-730 & 730nm & 10 Degrees & UK \B \\
\hline
\end{tabular}
\captionsetup[table]{skip=7pt}
\captionof{table}{List of the cameras, IR and lens setup used in this work.}
\label{tab:equipment}\vspace{-0.5cm}
\end{table*}

The primary objective of this paper is to not just create an object detector that could capture phone usage violations, but one that could do this during all hours of the day. In this section, we address the following challenges and propose solutions:

\subsubsection{Windscreen Glare}
\label{sec:approach:hardware:glare}

 \begin{figure}[htp]
    \centering
    \includegraphics[width=.4\textwidth]{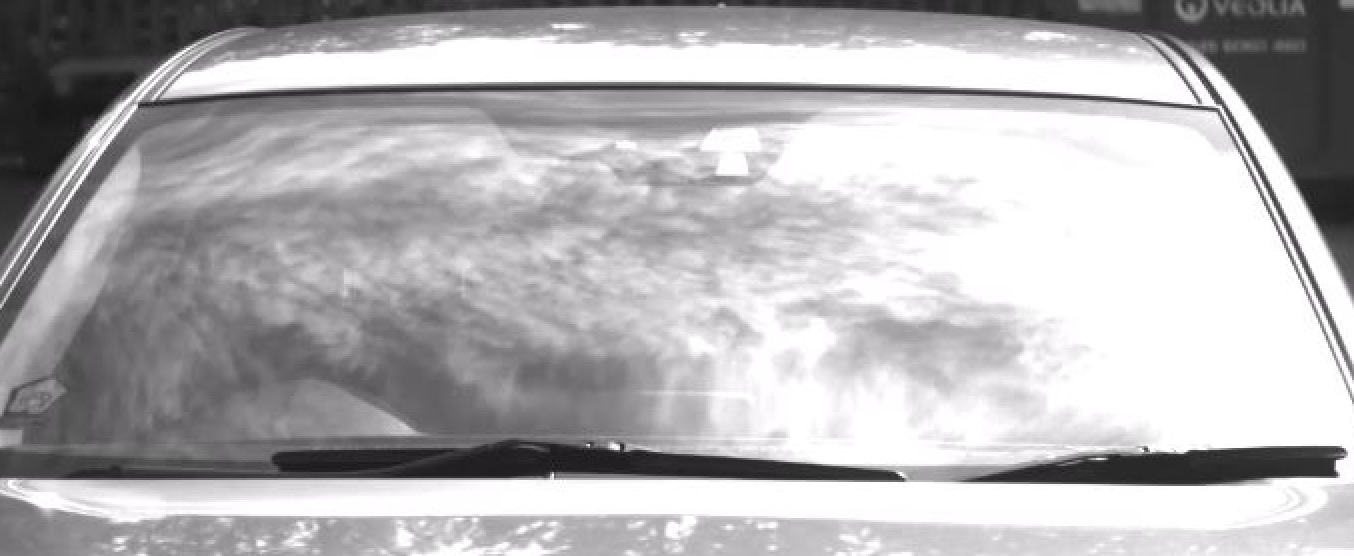}
    \caption{Image during sunny/cloudy day without a polarising filter to show the adverse effects of windscreen reflection.}
    \label{fig:jai}
\end{figure}

One of the most difficult challenges when trying to see inside the vehicle is windscreen glare. This will usually occur when the sun is in a particular part of the sky and can be made worse when clouds are present as they can be reflected quite significantly on the windscreen. An example of how glare and cloud reflection can completely obstruct the view into the vehicle from the windscreen can be seen in Figure \ref{fig:jai}. Our preliminary tests found that the issue of glare would occur during the majority of the day.

\begin{figure}
  \begin{subfigure}[t]{.5\textwidth}
    \centering
    \includegraphics[width=.8\textwidth]{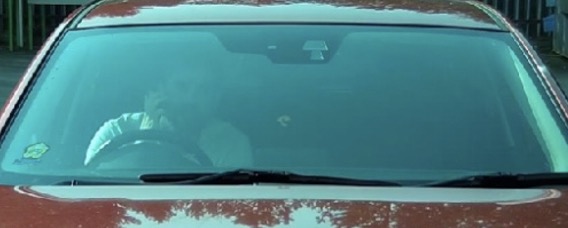}
    \caption{Without polarizing filter.}
  \end{subfigure}
  \hfill
  \begin{subfigure}[t]{.5\textwidth}
    \centering
    \includegraphics[width=.8\textwidth]{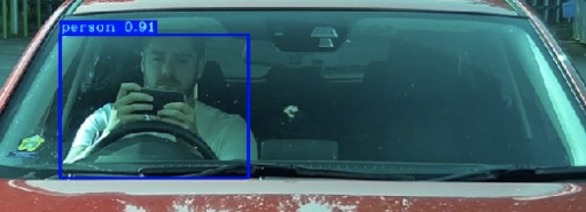}
    \caption{With polarizing filter.}
  \end{subfigure}
      \caption{Images passed through YOLOv3-416 object detector to detect a person. Top image without a polarizing filter is completely unable to detect the person. Bottom image with the polarizing filter is detecting the person with 91\% confidence.}
    \label{fig:polar1}\vspace{-0.4cm}
\end{figure}

We can solve this issue by using a polarising filter which is fixed to the camera lens. The effectiveness of this solution is demonstrated in Figure \ref{fig:polar1}, which shows the same image taken both without and with a polarising filter. We tested the impact of the polarising filter on the overall system by running the images through a pre-trained YOLOv3 model to see if it can detect the person inside the vehicle. We see from the images that it cannot detect the person when no filter is used, but detects the person with 91\% confidence when the polarising filter is added, which points to the importance of including a polarising filter with the system hardware.

\subsubsection{Low-light conditions}

\begin{figure}
  \begin{subfigure}[t]{.5\textwidth}
    \centering
    \includegraphics[width=.8\textwidth]{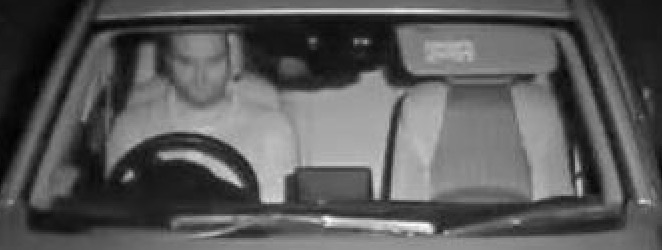}
    \caption{850nm IR.}
  \end{subfigure}
  \hfill
  \begin{subfigure}[t]{.5\textwidth}
    \centering
    \includegraphics[width=.8\textwidth]{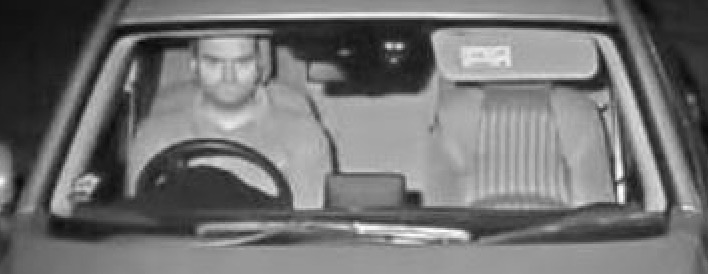}
    \caption{730nm IR.}
  \end{subfigure}
      \caption{Night images with active IR - 850nm vs 730nm.}
    \label{fig:IR}\vspace{-0.5cm}
\end{figure}

In order for the system to function successfully in low-light conditions, we would need to consider an appropriate light source. Directional white visible light is not used with the camera as it is considered to be a risk of causing glare to the driver. Instead, active Infra Red (IR)\footnote{\url{https://www.rayteccctv.com}} is used, where two different wavelengths are tested: 850 and 730 nanometers (nm). Figure \ref{fig:IR} demonstrates both wavelengths of IR working well in low-light conditions, however it is clear that the camera is able to capture more details using 730nm. Details of the IR used for this work can be found in Table \ref{tab:equipment}.

It is noted that IR can also be used during the day with the camera in monochrome setting, though more IR is required due to the higher ambient light levels. Further studies would be useful in determining the optimal IR power, typically measured in $\mu$W/cm$^2$, but this is outside the scope of this work.

The deployed system will be designed to expect both RGB (day) and monochrome images (night). A proportion of monochrome images taken with IR have been used in the training and testing of the model.

\subsection{Dataset}
\label{sec:appraoch:data}

This section describes the custom dataset used to train and test the models used in the proposed system.

\subsubsection{Training images}

\begin{figure*}[htp]
    \centering
    \includegraphics[width=2.5cm, height=2.5cm]{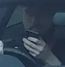}
    \hskip 0ex
    \includegraphics[width=2.5cm, height=2.5cm]{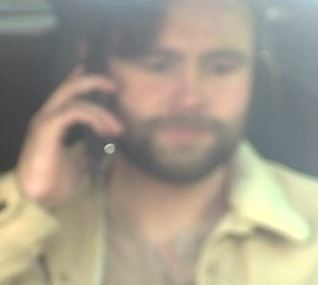}
    \hskip 0ex
    \includegraphics[width=2.5cm, height=2.5cm]{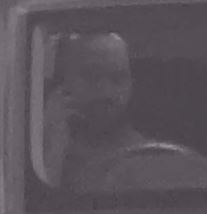}
    \hskip 0ex
    \includegraphics[width=2.5cm, height=2.5cm]{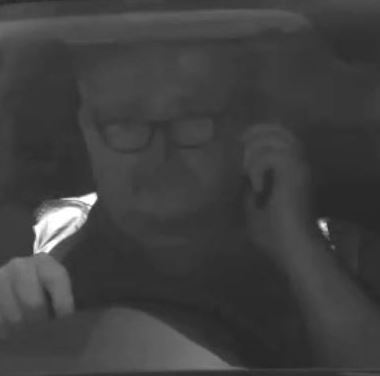}
    \hskip 0ex
    \includegraphics[width=2.5cm, height=2.5cm]{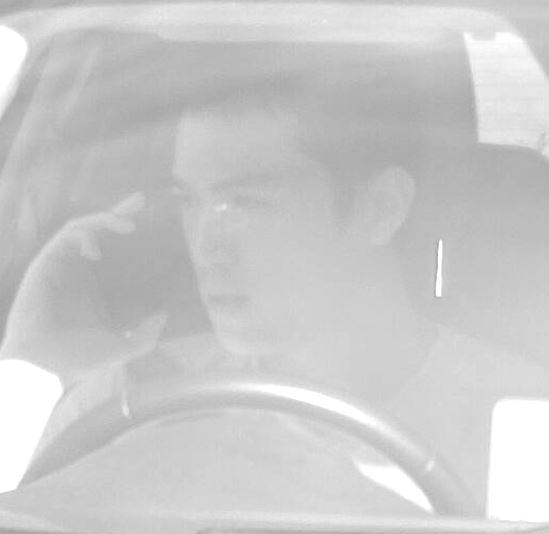}
    \hskip 0ex
    \includegraphics[width=2.5cm, height=2.5cm]{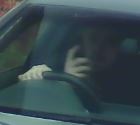}
    \hskip 0ex
    \caption{Bespoke RGB and monochrome training images of varying quality obtained specifically for this work.}\vspace{-0.5cm}
    \label{fig:images1}
    
\end{figure*}
For the single-step approach, the dataset consists of 2,150 images of phone and 2,235 images of licence plates. The licence plate images are obtained from the Google Open Images Dataset \cite{r33}. For the phone class, we used a small portion of mobile phone stock images in order to pick up simple detections. However, the main proportion of images would need to be obtained/created specifically for this project with a mixture of quality, weather conditions, and distance to best represent the real-world scenarios (Figure \ref{fig:images1}). Despite the class being labeled \enquote{phone}, the majority of what is actually being detected is multiple variations of hand positions holding a phone. In order to ensure that our model is trained to detect these types of images, 1,700 of the 2,150 phone images are obtained specifically for this project. 

The two-step approach required obtaining 263 images of vehicle windscreens, with some of these images having multiple vehicles, so the total number of windscreens annotated is 487. These images are used to train the first model in the two-step pipeline, which will be used to crop the vehicle windscreens. The second model is trained using only the phone images.

\subsubsection{Test images}
The images used to evaluate the trained models consist of 216 images of a person using their phone whilst driving. It is important to note that due to there being no access to public traffic camera footage, the test images were taken with the aid of volunteers to best represent that of a real-world application. These were obtained using a mixture of high-end and low-end cameras (Table \ref{tab:equipment}). The videos used to obtain these test images were not used in the training of the object detectors. Test images for the two-step approach were obtained by cropping out only the windscreen of these same images.

\subsection{Training the object detectors}
One of the main challenges for successfully detecting a person using their phone is the ability for the system to detect small objects with significant variation. To address this, multiple object detection methods are trained and evaluated. The same trained models are used to evaluate both the single-step and two-step approaches.

A variety of pre-trained base networks (backbones) can be chosen depending on the object detector used, such as ResNet \cite{r25}, VGG16 \cite{r12}, Inception \cite{r26} and MobileNet \cite{r27}. YOLO typically uses Darknet53 (YOLOv3) \cite{r9} and CSPDarknet53 (YOLOv4) \cite{r10}. All these base networks with the exception of MobileNet, are typically used when running on a GPU platform \cite{r20}. Here, we choose to include a MobileNet base network as one of our trained models to see how effective a low-cost light-weight detector could preform. An application such as this may benefit from running on an edge device \cite{r34} where a more light-weight model optimised for smaller computational resources would be preferred.

%  \begin{figure}[htp]
%     \centering
%     \includegraphics[width=12cm, height=4cm]{object detector.jpg}
%     \caption{Structure of an Object Detector \cite{r10}.}
%     \label{fig:obdet}
% \end{figure}

\subsubsection{YOLOv3 and YOLOv4}
Results from other studies \cite{r9, r10} concluded that the accuracy on the higher resolutions would yield greater accuracy, and similarly that the lower resolution models would run at a higher frame rate. Based on this, we opt for training both YOLOv3 and YOLOv4 with input resolutions of 512$\times$512, 416$\times$416 and 320$\times$320. For the two-step approach, we would like to see if accuracy would be impacted much on the lower resolution models dealing with: uniform objects such as windscreens, and low resolution cropped images taken from the windscreen.

\subsubsection{Faster R-CNN, SSD, and Centernet}

The remaining frameworks are obtained and fine-tuned using the TensorFlow Object Detection API \cite{r15, r16}, this give us access to many different pre-trained models that could be fine-tune on our custom dataset.

\subsection{Evaluating the models}

To replicate a real-world scenario, the majority of the test images are obtained at a distance of 20-30m from the camera, with a height of approximately 3m. These images were captured during different times of the day under varying weather conditions to enable testing the generalisation capabilities of the system as well as its predictive performance.

Test images for evaluation are split into 2 sets, the first using 216 full image snapshots taken from the camera, the next with the same images but this time with only the cropped windscreen. This allows us to determine if the model will perform more favourably with the single-step or the two-step system.

\subsubsection{Evaluation Metrics}

 \begin{figure}[htp]
    \centering
    \includegraphics[width=.4\textwidth]{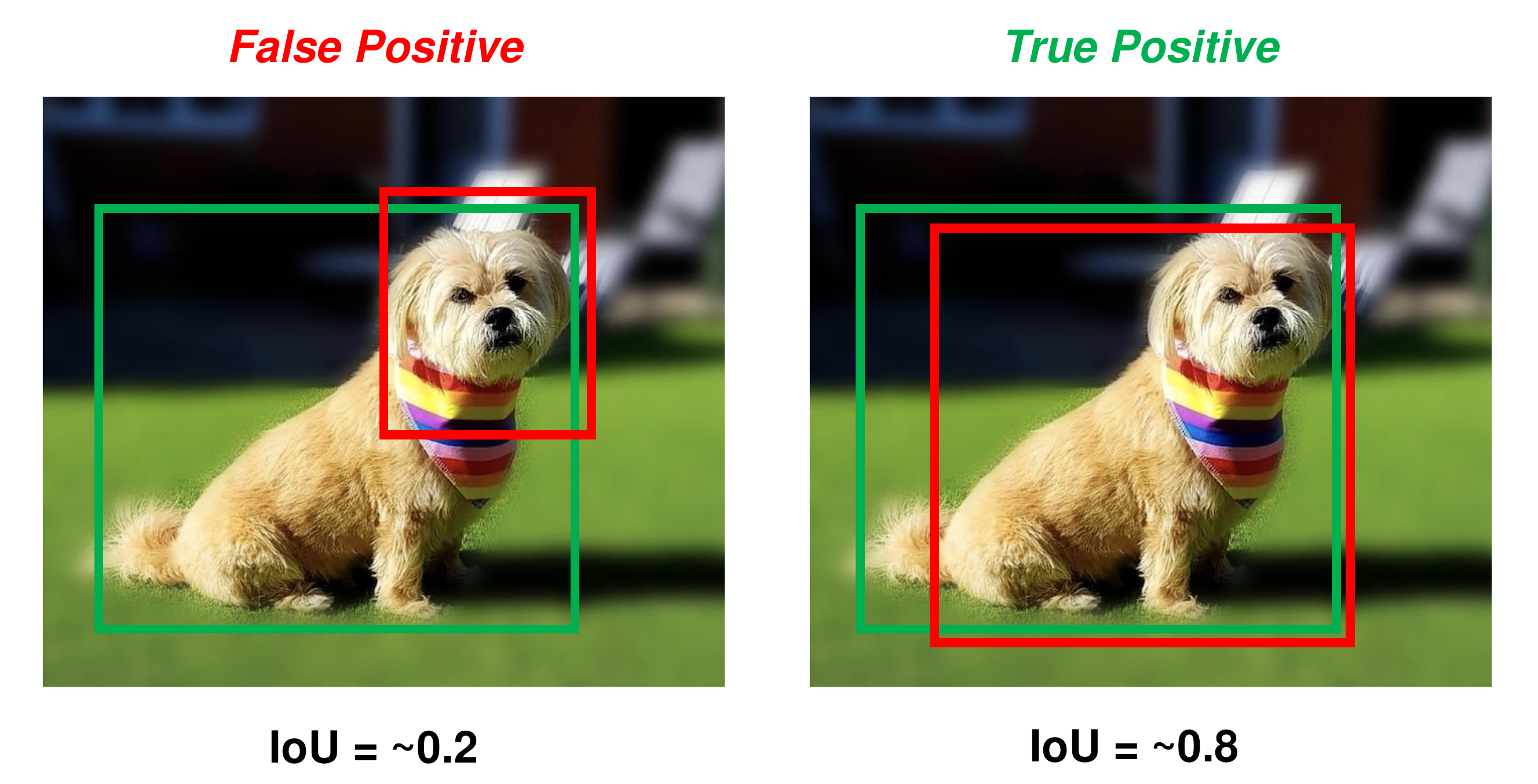}
    \caption{Example of False Positive (left) and True Positive (right) when IoU threshold set to $>$0.5. Green bounding box represents the ground truth, red bounding box is the prediction.}
    \label{fig:iou}\vspace{-0.6cm}
\end{figure}

A good metric to evaluate object detectors is Mean Average Precision (mAP) \cite{r11}. However, as we are only concerned with the accuracy of a single class (Phone), we evaluate the approach using Average Precision (AP) \cite{r30}. The first part of this process is collecting the sequences of True Positives (TP) and False Positives (FP) from the predictions made on the test images. In object detection, a TP is determined by the value of the Intersection Over Union (IoU). For example, if the minimum IoU requirement is $>$0.5 (often referred to as $mAP_{50}$ or $AP_{50}$, where any predictions with IoU above this threshold are classed as TP and any below are FP \cite{r28} (Figure \ref{fig:iou}). Once these have been collected, they are sorted in descending order by the confidence score, then precision and recall values are calculated using Equation \ref{eqn:pr}. 

%\vspace{-0.5cm}
\begin{equation}
\setlength{\abovedisplayskip}{1pt}
\setlength{\belowdisplayskip}{4pt}
%\begin{split}
Precision =   \frac{TP}{TP+FP} 
%   \quad\mathrm{and}\quad \\
\; \; \; Recall =  \frac{TP}{TP+FN}%(Ground truth)
\label{eqn:pr}
%\end{split}
\end{equation}

The AP summarises the shape of the precision/recall curve, and is defined as the mean precision at a set of eleven equally spaced recall levels [0, 0.1,..., 1] \cite{r30} (Equation \ref{eqn:AP}). The precision at each recall level $r$ is interpolated by taking the maximum precision measured for a method for which the corresponding recall exceeds $r$ \cite{r30} (Equation \ref{eqn:AP}).

In this work, we evaluate our proposed approach using both IoU$>$0.5 and IoU$>$0.1, hereby referred to as $AP_{50}$ and $AP_{10}$ respectively. We are attempting to find very small objects (mobile phone) with significant variation, so the lower IoU threshold would likely be more appropriate. To test the accuracy of our trained models, all the test images are supplied with given annotation files (ground truth bounding boxes). We then run these images through the object detectors to obtain the predicted bounding boxes. The AP is subsequently calculated based on both IoU thresholds. Next, we discuss evaluating the efficiency of the proposed approach.

\begin{equation}
\setlength{\abovedisplayskip}{1pt}
\setlength{\belowdisplayskip}{3pt}
\begin{split}
& AP = \frac{1}{11} \sum\limits_{r\in \{0,0.1,0..,0.9,1\}} Pinterp^{(r)}
\\ & \quad\mathrm{where}\quad
Pinterp^{(r)}= \max\limits_{\tilde{r}:\tilde{r}\le r} p(\tilde{r})\\
\label{eqn:AP}
\end{split}
\end{equation}\vspace{-0.5cm}

\subsubsection{Evaluating frame rate}

The system has been designed around the ability to take live video from traffic cameras. It is therefore important that it not only detects with high accuracy, but with low latency, especially when dealing with moving vehicles. Other studies confirm that AP alone is not enough to evaluate object detectors, particularly when it comes to \emph{video} object detection \cite{r24}. Consequently, frames per second (FPS) is another metric used to evaluate the proposed system. To test speed of our trained object detectors, we run the same test video for each model and then calculate the average FPS.

\subsubsection{Choosing the best model}

Once these metrics are calculated for each of the models, we then shortlist the top two for further evaluation. The test images are split into two categories, namely high-quality and low-quality. High-quality images are captured using the high-end cameras (Avigilon and Axis) to represent how the model should perform when the system has been built with a relatively larger budget in mind. The low-quality images are acquired using the low-end camera (ELP) to demonstrate how the model will perform under cost-effective considerations. Details of the cameras are listed in Table \ref{tab:equipment}. The speed in which these models perform also needs to be re-evaluated to incorporate the two-step windscreen method as well as the object tracking algorithm.

\subsubsection{Object tracking and data collection}

A consideration when building a fully-automated system is how the phone violations are going to be recorded in a way that is useful to the end user. To do this, the system would have to be able to distinguish between unique and duplicate detection. For example, a five second video may show one driver using their phone, but since the detections are done per frame, it may count duplicate violations for every one of these frames. In order to address this, we add DeepSort \cite{r8}, which is an object tracking algorithm. This will add a unique ID for each detection and then takes each frame to predict if the next detection belongs to the same ID or not. In case of the single-step system, we can check every new detection to see if the ID has been seen before, then if not, can log as a new detection. For the two-step system, a phone violation is only logged once per unique windscreen ID. This same method also allows us to count the number of vehicles, taken from the licence plate on the single-step method and windscreen on the two-step.

\section{Experimental Results}
\label{sec:results}

In this section, we evaluate our models using the experimental setup and the metrics discussed in the previous section.

\subsection{System Specification}

Trained object detectors are tested and evaluated using the system specification listed in Table \ref{tab:system}.

\begin{table}[h]
\centering
\begin{tabular}{ p{4cm}p{3cm} }
\hline
\rowcolor{lightgray}Type & Spec\T\B \\
\hline
Processor & AMD Ryzen 7 3800X\T  \\
GPU & Nvidia RTX2080Ti   \\
Memory & 32GB \\
Operating System    & Windows 10  \\
Programming Language & Python 3.8\\
Machine Learning Platform & TensorFlow 2.2\B \\
\hline
\end{tabular}
\captionsetup[table]{skip=7pt}
\captionof{table}{System Specifications.}
\label{tab:system}\vspace{-0.4cm}
\end{table}

\subsection{Average precision}

\begin{table*}[h]
\centering
\begin{tabular}{ p{4.3cm}p{1.8cm}p{1.8cm}p{1.8cm}p{1.8cm}p{1.8cm} }
\hline
\rowcolor{lightgray}Object detector & $AP_{50}$ & $AP_{10}$ & $AP_{50}$ cropped & $AP_{10}$ cropped & FPS (detection)\T\B \\
\hline
YOLOv4 512 & \textbf{45.98} & \textbf{73.11} & 59.62 & 83.32 & 27.12\T  \\
YOLOv3 512 & 35.81 & 46.16 & \textbf{63.27} & 85.88 & 25.18 \\
YOLOv4 416 & 40.23 & 48.23 & 61.60 & \textbf{87.58} & 26.12 \\
YOLOv3 416 & 37.41  & 51.64 & 58.44 & 79.99 & 25.96 \\
YOLOv4 320 & 19.62 & 52.36 & 52.65 & 81.93 & 26.05 \\
YOLOv3 320 & 37.54 & 72.45 & 59.05 & 84.62 & 28.97 \\
Centernet ResNet101 512 & 43.81 & 56.18 & 50.04 & 66.6 & 35.9 \\
Faster R-CNN ResNet101 640 & 37.18 & 48.57 & 41.78 & 55.5 & 15.4 \\
Faster R-CNN ResNet152 640 & 34.85 & 44.48 & 40.89 & 50.57 & 11.5 \\
SSD Mobilenet FPNLite 640 & 12.23 & 18.23 & 12.64 & 28.41 & \textbf{42.77} \\
SSD ResNet50 FPN 640 & 2.77 & 7.77 & 0.92 & 4.22 & 23.66 \\
SSD ResNet101 FPN 640 & 0.6 & 4.86 & 0.71 & 13.9 & 22.2\B \\
\hline
\end{tabular}
\captionsetup[table]{skip=7pt}
\captionof{table}{Results of all the trained models showing average precision and frames per second. Cropped refers to images of the windscreen only to evaluate performance of the two-step approach. the frame per second (FPS) metric is based on detection only and does not include tracking and the two-step approach.}
\label{tab:results}
\end{table*}

\begin{figure*}
  \begin{subfigure}[t]{.5\textwidth}
    \centering
    \includegraphics[width=.8\textwidth]{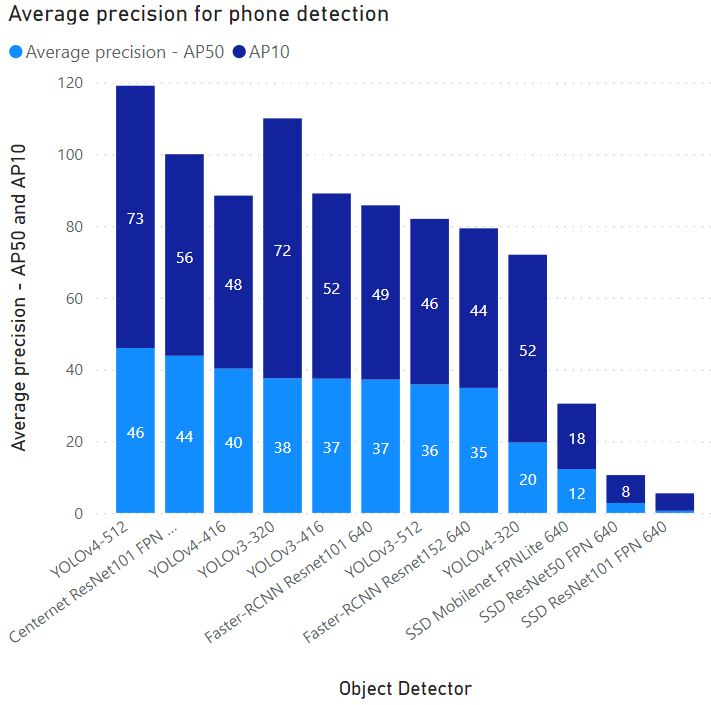}
    \caption{Single-step approach.}
  \end{subfigure}
  \hfill
  \begin{subfigure}[t]{.5\textwidth}
    \centering
    \includegraphics[width=.8\textwidth]{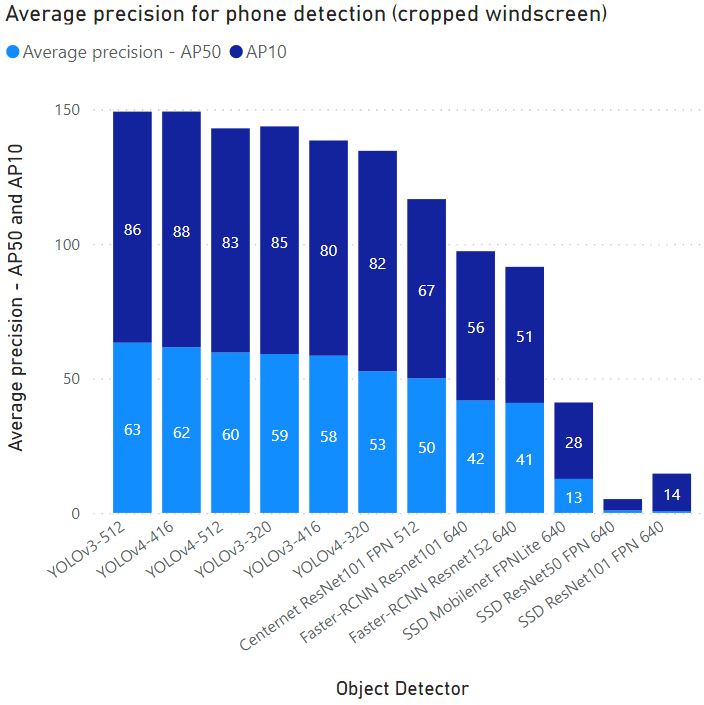}
    \caption{Two-step approach.}
  \end{subfigure}
      \caption{Average precision for both IoU thresholds of 0.5 and 0.1 - single-step (left) \& two-step (right).}
    \label{fig:bar_results_comb}\vspace{-0.5cm}
\end{figure*}

For the first accuracy test, we present the single-step method where the full test image is used in the model trained to detect the phone. Figure \ref{fig:bar_results_comb} shows the YOLO models outperforming the other object detectors. YOLOv4 with input size 512 is best performing on both $AP_{50}$ and  $AP_{10}$, whilst the SSD models are the poorest performing for both IOU thresholds.

%  \begin{figure}[htp]
%     \centering
%     \includegraphics[width=.3\textwidth]{fp.jpg}
%     \caption{False positive result for $AP_{50}$.}
%     \label{fig:fp}
% \end{figure}

Next, we test the same models, but this time with the cropped windscreen images which will allow us to determine whether the two-step approach is more appropriate. Figure \ref{fig:bar_results_comb} suggests that if accuracy was the main driver, the two-step method will be more favourable to use, giving higher AP in almost all of the trained object detectors. Again, the YOLO models yield the highest accuracy scores. YOLOv3 with the larger input size gives the best results with $AP_{50}$, whilst YOLOv4 with the input resolution of 416 gives the highest accuracy for $AP_{10}$. Once again, accuracy for all 3 SSD frameworks are the lowest by a significant margin. Another study \cite{r20} points out that, in SSD object detectors, regardless of which base network you use, it will still retain the original characteristics of the SSD. Therefore, accuracy on small objects will not be as accurate compared to other two-stage models such as R-CNN.

Review of the predictions made on the test images when IOU threshold is set to $>$0.5 shows multiple false positives despite the predicted bounding box surrounding the correct object. As mentioned previously, the objects that the model is attempting to predict have significant amounts of variation, meaning that it will always be difficult to get a high IOU score. Figure \ref{fig:fp} shows a false positive prediction where we have the IOU threshold set to $>$0.5. We can see that the model is capturing the violation correctly, but narrowly missing the IOU threshold resulting in a false positive. Based on this, we propose an IOU threshold of $>$0.1 for this application.

\subsection{Frame rate}
Speed of the trained models is evaluated using a 70-second test video. These initial frame rate evaluation tests are done on detection only prior to adding the tracking algorithm and the two-step approach.

\begin{figure}[htp]
    \centering
    \includegraphics[width=.4\textwidth]{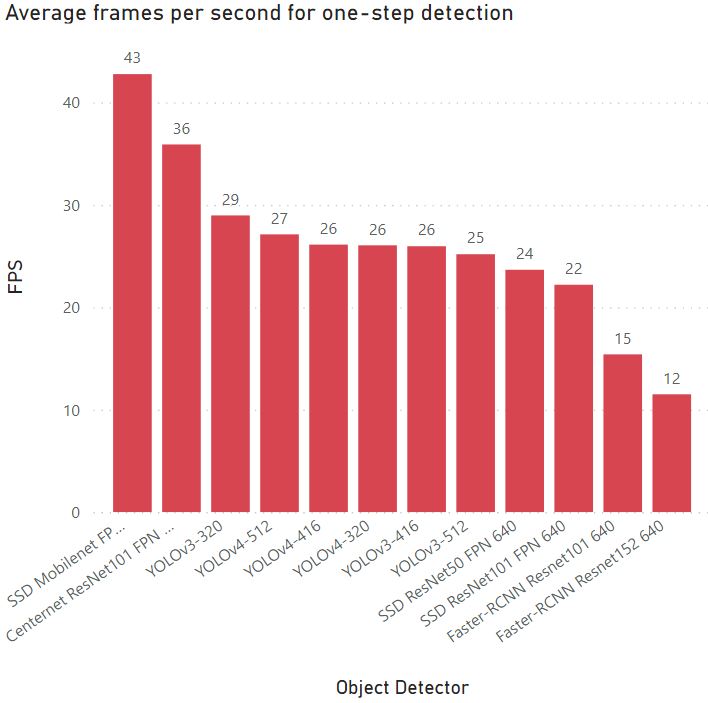}
    \caption{FPS for detection only for the trained object detectors.}
    \label{fig:fps}
\end{figure}

As seen in Figure \ref{fig:fps}, the single-stage (YOLO, SSD, CenterNet) object detectors are significantly faster than the two-stage R-CNN detectors. As expected, the most efficient object detector is the low-cost SSD Mobilenet FPNLite 640 with 43 FPS, however lacking in accuracy. Each of the YOLO models seem to perform consistently well with regards to both accuracy and speed with YOLOv3 320 performing the best at 29 FPS.

\subsection{Output images}

\begin{figure}[!tbp]
  \centering
  \subfloat{\includegraphics[width=3cm, height=2.7cm]{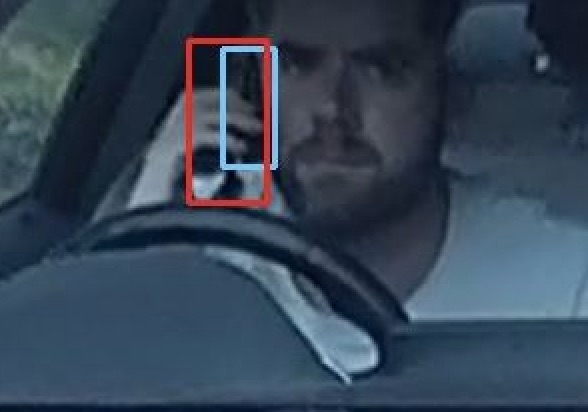}}
  \hskip 1ex
  \subfloat{\includegraphics[width=3cm, height=2.7cm]{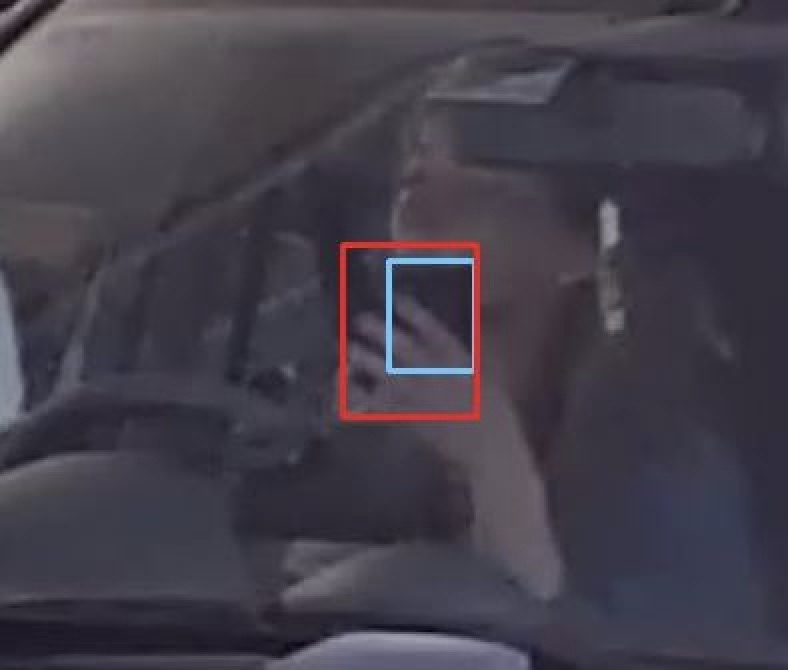}}
  \caption{False positive result for $AP_{50}$, which demonstrates that for the application in this work, $AP_{10}$ is more appropriate.}
  \label{fig:fp}\vspace{-0.5cm}
\end{figure}

\begin{figure} 
\centering
\begin{tabular}{*{3}{@{}c}@{}}
\includegraphics[width=2.5cm, height=2.5cm]{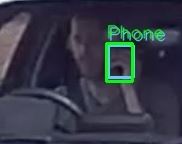}    & 
\hskip 1ex
\includegraphics[width=2.5cm, height=2.5cm]{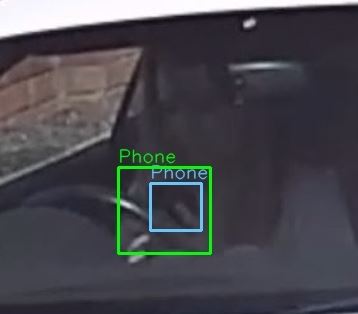}   & 
\hskip 1ex
\includegraphics[width=2.5cm, height=2.5cm]{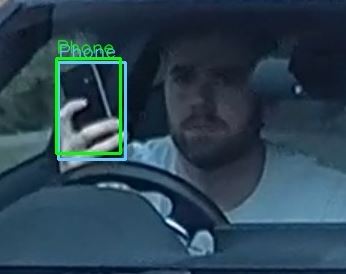}  \\
\includegraphics[width=2.5cm, height=2.5cm]{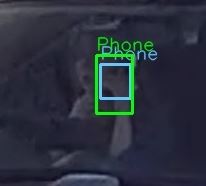}   & 
\hskip 1ex
\includegraphics[width=2.5cm, height=2.5cm]{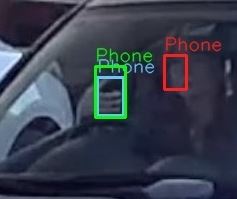}  & 
\hskip 1ex
\includegraphics[width=2.5cm, height=2.5cm]{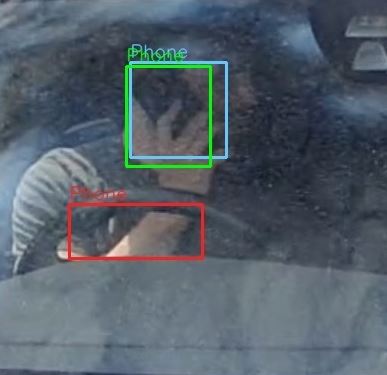}  
\end{tabular}
\caption{Results from YOLOv3 320 with $AP_{10}$ showing high accuracy but high false positives on low-quality images. Blue bounding boxes denote ground truth, green refers to the true positive prediction and red is the false positive perdition.}
\label{fig:detections}\vspace{-0.25cm}
\end{figure}

Figure \ref{fig:detections} shows a number of sample results obtained from the YOLOv3-320 model for the two-step method; blue bounding box refers to the ground truth, green is the true positive prediction, and red is the false positive perdition. With regards to predictions made on the low-quality camera, although the model performs well, there appears to be a higher chance of false positive predictions.

Having a high proportion of false positives for this particular application could result in incorrectly fining individuals, or involve human intervention to manually go through all the violations, which is both costly and defeats the main purpose of this work, which is to automate the pipeline. For the final deployed model, it may be appropriate to increase the score threshold for the phone detector step to reduce these false positives. Observation from the predictions of the test set confirm that $AP_{10}$ is appropriate for this application of detecting such small and difficult images, as demonstrated in Figure \ref{fig:fp}.

Based on the results shown in Table \ref{tab:results}, the model chosen for deployment is YOLOv3 with the input size of 320 using the two-step method. Although it did not achieve the highest overall accuracy, it came a close third behind YOLOv3-512 and YOLOv4-416. The deciding factor is the speed of the model, as it is able to achieve almost 29 FPS (almost 11\% faster then the next best performing model). For this type of application, the cameras typically monitors fast-moving traffic. The model should consequently be able to make detections efficiently. Having a model with a smaller input size means it will be less expensive with regards to hardware demands.

\subsection{Integrating the two-step method and tracking}

The next stage in our overall system is to re-train the chosen YOLOv3-320 model and modify the code for the two-step approach, where first step is trained to detect the windscreen and the next step trained on only the phone images. The DeepSort \cite{r8} tracking algorithm is also integrated into this system. The frame rate is recalculated on the same video, to show the impact of the tracking algorithm and the two-step approach on system efficiency. Table \ref{tab:results3} shows that the tracking algorithm reduces FPS by almost 10\%, whilst adding the extra step on top of this sees a further reduction of $\sim$50\% giving a frame rate of 13.15 FPS on the YOLOv3-320 model.

\begin{table*}[h]
\centering
\begin{tabular}{ p{2.5cm}p{2.8cm}p{2.8cm}p{3.8cm}}
\hline
\rowcolor{lightgray}Object detector & FPS & FPS & FPS\T    \\
\rowcolor{lightgray}   & (detection only) & (with tracking) & (with tracking \& two-step)\B \\
\hline
YOLOv3 320 & 29.71 & 26.94 & 13.15\T \\
\hline
\end{tabular}
\captionsetup[table]{skip=7pt}
\captionof{table}{Impact on frame rate for the chosen model YOLOv3-320 when tracking and two-step is added. Frame rate of the two-step has been recorded when there is constant activity in the video. Without activity, frame rate increases to that of detection and tracker.}
\label{tab:results3}\vspace{-0.22cm}
\end{table*}

\begin{table*}[h]
\centering
\begin{tabular}{ p{2.5cm}p{2.1cm}p{2.1cm}p{2.1cm}p{2.1cm}p{2cm}p{2cm}p{2cm} }
\hline
\rowcolor{lightgray}Object detector & Model type &$AP_{50}$  & $AP_{10}$ & $AP_{50}$  & $AP_{10}$\T    \\
\rowcolor{lightgray} & & (high-quality) & (high-quality) &(low-quality) & (low-quality)\B  \\
\hline
YOLOv3 320 & Two-step & 78.29 & 95.81 & 41.93 & 74.36 \T  \\
\hline
\end{tabular}
\captionsetup[table]{skip=7pt}
\captionof{table}{Average precision results when images are split between high-end and low-end cameras for YOLOv3-320.}
\label{tab:results2}\vspace{-0.5cm}
\end{table*}

For the final benchmark tests, we split the test images into 2 categories; high-quality and low-quality, with 116 and 100 images respectively. Based on our chosen metric of IOU threshold greater than 0.1, for YOLOv3-320, we can achieve an AP of as high as 95.81\% on the images taken with only high-quality cameras, whilst still achieving an AP of 74.36\% on images taken from the low-quality camera (Table \ref{tab:results2}). 

\section{Discussions and future work}
\label{sec:discussions}

Our proposed approach delivers very promising results and further enables a fully-automated end-to-end surveillance system capable of capturing mobile use violations while driving. However, there are still limitations that need to be addressed before tangible impact can be made.

The proposed two-step model detects the driver side of the windscreen based on right-hand drive vehicles. When deployed in countries using left-hand drive vehicles, we can simply crop the opposite side. Alternatively, even this process can be automated by detecting the licence plate to identify country and determine which side is the driver.

Section \ref{sec:appraoch:data} addresses not having access to public roadside cameras, so next step would be to deploy the system with support from local authority/police. Test parameters of this work have been based on 3m mounting height with a distance of 25-30m from the subject, meaning that phone could be hidden when texting close to lap. This could be remedied when deployed on a public road by utilizing a gantry traffic camera which allows for a better view within the vehicle.

Although, the two-stage approach achieves a greater accuracy, there is a compromise with frame rate. We propose optimizing the model and exploring TensorRT \cite{r39} framework to potentially improve the speed of the model.

\section{Conclusion}
\label{sec:conclusion}

In this paper, we have presented a deep learning approach for detecting driver phone violations in all weather conditions without the need for human intervention. A total of 12 object detection models \cite{r4, r10, r19, r25, r21, r20, r27} are fine-tuned and evaluated based on speed and accuracy for both the approaches which are: single-step, where a single frame is used to detect the phone, and the two-step, which first detects windscreen and then uses the cropped image of only the driver side to detect the phone. The two-step approach yields higher accuracy but lower frame rate due to having to run two models simultaneously. The model chosen based on both accuracy and speed is YOLOv3 with an input resolution of 320. We also integrate DeepSort, an object tracking algorithm, which allows us to only collect and log unique phone detections from the driver's side, meaning that this collected data could be made useful for time-series analysis. The trained object detector is able to achieve an accuracy of 84.62\% ($AP_{10}$) on the 216 test images, with a frame rate of 13.15 FPS during activity and $\sim$27 with no activity. When images were split between high-end camera and low-end camera, we achieve accuracy levels as high as 95.81\% on the high-end and 74.36\% on the budget cameras. A user interface is also built in order for the user to easily access the information as well as the ability to view snapshots of all violations.

\textit{We kindly invite the readers to refer to the supplemental \href{https://youtu.be/PErIUr3Cxvg}{\textbf{video}}: \url{https://youtu.be/PErIUr3Cxvg} for more information and more results in video format.}

\bibliographystyle{unsrt}  
\bibliography{references}

\end{document}